\documentclass[11pt,a4paper,fleqn]{article}

\usepackage[margin=1in]{geometry}
\usepackage[T1]{fontenc}
\usepackage{lmodern}
\usepackage{microtype}
\usepackage{amsmath,amssymb}
\usepackage{graphicx}
\usepackage{booktabs}
\usepackage{multirow}
\usepackage{adjustbox}
\usepackage{svg}
\usepackage{placeins}
\usepackage[authoryear]{natbib}
\usepackage[hidelinks]{hyperref}
\usepackage{xparse}
\usepackage{etoolbox}
\usepackage{pgfkeys}

\makeatletter

\newcommand{\preprint@title}{}
\newcommand{\preprint@authors}{}
\newcommand{\preprint@email}{}

\newcounter{preprintauthorcount}
\newif\ifpreprint@frontprinted
\preprint@frontprintedfalse

\NewDocumentCommand{\shorttitle}{m}{}
\NewDocumentCommand{\shortauthors}{m}{}

\RenewDocumentCommand{\title}{o m}{%
  \gdef\preprint@title{#2}%
}

\RenewDocumentCommand{\author}{o m o}{%
  \ifnum\value{preprintauthorcount}=0
    \IfNoValueTF{#1}{%
      \gdef\preprint@authors{#2}%
    }{%
      \ifstrempty{#1}{%
        \gdef\preprint@authors{#2}%
      }{%
        \gdef\preprint@authors{#2\textsuperscript{#1}}%
      }%
    }%
  \else
    \IfNoValueTF{#1}{%
      \gappto\preprint@authors{, #2}%
    }{%
      \ifstrempty{#1}{%
        \gappto\preprint@authors{, #2}%
      }{%
        \gappto\preprint@authors{, #2\textsuperscript{#1}}%
      }%
    }%
  \fi
  \stepcounter{preprintauthorcount}%
}

\NewDocumentCommand{\cormark}{o}{}

\NewDocumentCommand{\ead}{m}{%
  \gdef\preprint@email{#1}%
}

\newcommand{\preprint@printfront}{%
  \ifpreprint@frontprinted
  \else
    \global\preprint@frontprintedtrue
    \begin{center}
      {\LARGE\bfseries \preprint@title\par}
      \vspace{0.75em}
      {\large \preprint@authors\par}

      \ifx\preprint@email\@empty
      \else
        \vspace{0.35em}
        {\normalsize\texttt{\preprint@email}\par}
      \fi
    \end{center}
    \vspace{0.5em}
  \fi
}

\pgfkeys{
  /preprintaff/.is family,
  /preprintaff,
  organization/.store in=\preprintafforg,
  addressline/.store in=\preprintaffaddr,
  city/.store in=\preprintaffcity,
  postcode/.store in=\preprintaffpost,
  country/.store in=\preprintaffcountry
}

\NewDocumentCommand{\affiliation}{o m}{%
  \preprint@printfront

  \def\preprintafforg{}%
  \def\preprintaffaddr{}%
  \def\preprintaffcity{}%
  \def\preprintaffpost{}%
  \def\preprintaffcountry{}%

  \pgfkeys{/preprintaff,#2}%

  \begin{center}
    \small
    \IfNoValueF{#1}{%
      \ifstrempty{#1}{}{\textsuperscript{#1}}%
    }%
    \textit{\preprintafforg}\\
    \preprintaffaddr, \preprintaffcity, \preprintaffpost, \preprintaffcountry
  \end{center}%
}

\NewDocumentCommand{\cortext}{o m}{%
  \preprint@printfront
  \par\noindent{\footnotesize #2}\par
}

\RenewDocumentCommand{\maketitle}{}{%
  \preprint@printfront
}

\NewDocumentEnvironment{highlights}{}{%
  \begin{quote}
    \begin{itemize}
      \setlength{\itemsep}{0.15em}
}{%
    \end{itemize}
  \end{quote}
}

\NewDocumentEnvironment{keywords}{}{%
  \par\smallskip
  \noindent\textbf{Keywords: }%
  \begingroup
}{%
  \endgroup
  \par\medskip
}

\ProvideDocumentCommand{\bio}{m}{}

\makeatother

\begin{document}
\let\WriteBookmarks\relax
\def\floatpagepagefraction{1}
\def\textpagefraction{.001}

\shorttitle{}    
\shortauthors{}  

% Title
\title [mode = title]{Brain-PACE: A Deep Siamese MRI Framework for Modelling Longitudinal Brain Acceleration}

\author[1]{Samuel Maddox}[orcid={0009-0007-0944-992X}]
\cormark[1]

\author[1]{Jacob Newman}

\author[2]{Saber Sami}

\author[1]{Michal Mackiewicz}

\author[]{for the Alzheimer's Disease Neuroimaging Initiative}
\cormark[2]

\author[]{the Australian Imaging Biomarkers and Lifestyle flagship study of ageing}
\cormark[3]

% Affiliations
\affiliation[1]{
organization={School of Computing Sciences, University of East Anglia},
addressline={Norwich Research Park},
city={Norwich},
postcode={NR4 7TJ},
country={UK}
}

\affiliation[2]{
organization={Norwich Medical School, University of East Anglia},
addressline={Norwich Research Park},
city={Norwich},
postcode={NR4 7TJ},
country={UK}
}

\cortext[2]{Data used in preparation of this article were obtained from the Alzheimer's Disease Neuroimaging Initiative (ADNI) database (adni.loni.usc.edu). As such, the investigators within the ADNI contributed to the design and implementation of ADNI and/or provided data but did not participate in the analysis or writing of this report. A complete listing of ADNI investigators can be found at: http://adni.loni.usc.edu/wp content/uploads/how\_to\_apply/ADNI\_Acknowledgement\_List.pdf. }

\cortext[3]{Data used in the preparation of this article was obtained from the Australian Imaging Biomarkers and Lifestyle flagship study of ageing (AIBL) funded by the Commonwealth Scientific and Industrial Research Organisation (CSIRO) which was made available at the ADNI database (www.loni.usc.edu/ADNI). The AIBL researchers contributed data but did not participate in analysis or writing of this report. AIBL researchers are listed at data.aibl.org.au/adni/.}

\begin{abstract}
Brain age estimation has become a popular research proxy for assessing brain health and disease, yet longitudinal trajectories of brain ageing are still poorly defined, and clinical use is limited. Building on existing Siamese longitudinal frameworks, we develop Brain-Predicted Age Acceleration (Brain-PACE) to directly estimate the pace of structural brain ageing from paired T1-weighted MRI. Brain-PACE identified accelerated ageing in 42.6\% of participants with mild cognitive impairment. Faster Brain-PACE was associated with greater functional and cognitive impairment (FAQ; $r=0.35$, ADAS13; $r=0.30$, CDR-SB; $r=0.32$) and greater regional tau burden in the posterior cingulate ($r=0.59$), precuneus ($r=0.47$), and entorhinal cortex ($r=0.37$). These associations were stronger than those observed when pace was calculated indirectly from repeated cross-sectional brain age estimates, suggesting that direct longitudinal modelling captures complementary information relevant to ongoing pathological change. Methodologically, Brain-PACE extends the LILAC framework by combining spatial attention with soft label distribution learning and a Cram\'er distance objective, improving probabilistic performance and reducing prediction bias while providing measures of predictive uncertainty. Together, these findings support Brain-PACE as a complementary longitudinal imaging phenotype with sensitivity to relevant clinical and biological changes in early neurodegeneration.
\end{abstract}

\maketitle

\newpage       
% Main text
\FloatBarrier 
\section{Introduction}\label{}

Ageing clocks have advanced rapidly in recent years and are increasingly being used to characterise age related molecular and physiological variation across diverse populations. Using machine learning, they estimate how old an individual's biology appears relative to a reference population. For example, epigenetic clocks have been associated with morbidity, mortality and functional decline \citep{faul2023epigenetic}, as well as recent work from \citet{kim2026epigenetic} showing links between accelerated epigenetic ageing and blood markers of neurodegeneration in older Indian adults. Alongside this, proteomic ageing clocks have shown ability to predict disease across geographically and genetically distinct populations \citep{argentieri2024proteomic}. These ageing clocks can also be presented across various biological scales. Organ-specific proteomic and large-scale MRI models demonstrate how different organ systems within the same individual can exhibit distinct profiles \citep{oh2023organ, multi2026mri}. Within neuroimaging, MRI derived brain age has become a commonly applied summary marker of global brain health and disease. It can be estimated from routine T1w structural MRI, with the derived brain Predicted Age Difference PAD (brain-PAD) representing the deviation between predicted and chronological age \citep{cole2017predicting, peng2021accurate}. Brain-PAD has been associated with many neurodegenerative conditions including Parkinson's disease and Alzheimer's disease (AD) \citep{hahnel2025padpark, lee2022deep}, making it a popular, accessible imaging-derived phenotype.

However, ageing is inherently a longitudinal process and most machine learning-derived measures of biological age are cross-sectional. Brain-PAD similarly only represents an individual's position relative to a population. This can be subject to baseline effects and does not necessarily quantify their current rate of person specific deviation \citep{vidal2021individual, smith2025characterising}. For example, an individual may consistently exhibit a positive brain-PAD across multiple time points without ongoing structural change. Furthermore, the utility of brain-PAD has shown limited benefit over more interpretable T1w MRI features \citep{dorfel2026prediction, korbmacher2025cross_hbm}. These limitations are particularly relevant for evaluating AD, where pathology develops over an extended preclinical window. Structural changes associated with healthy ageing can also overlap with those observed during early AD stages \citep{fjell2014normal}. While Brain-PAD can highlight clinically relevant information \citep{zhang2025brain}, to better characterise brain ageing in these early disease stages, there is a need to move from evaluating whether a brain looks older than expected at a single time point towards measuring how quickly it is changing between time points. For example, longitudinal changes in established epigenetic clocks have been related to subsequent mortality \citep{kuo2026longitudinal}, and recent proteomic work derived organ-specific pace from changes between independently estimated organ ages \citep{liu2026accelerated}. These studies demonstrate the value of incorporating temporal information, but the underlying ageing models are inherently not optimised to estimate longitudinal change of biological state. An important distinction therefore exists between directly estimating longitudinal ageing and calculating it indirectly from repeated cross-sectional predictions. The latter remains dependent on two independent estimates which may propagate prediction error from each time point. In contrast, longitudinal models directly learn the differences occurring within the same participant. Whether this provides biologically meaningful information beyond conventional methods remains unclear.

Longitudinal modelling approaches are still in the early stages of addressing this shift. \citet{ouyang2022self} shows how a longitudinal neighbourhood embedding approach can produce globally consistent progression trajectories that represent the rate of brain ageing. More recently, the Learning-based Inference of Longitudinal imAge Changes (LILAC) approach has been developed for temporal difference prediction \citep{kim2025pace2}. This framework was further optimised in the following LILAC+ architecture which replaced the linear head with a Multi-Layer Perceptron (MLP) \citep{wegmann2025pace3}. The MLP showed improved ability in modelling pace, potentially by capturing the non-linear relationships present in patterns of human ageing \citep{mousley2025topological}. Other recent studies from \citet{yin2025pace1, whitman2025dunedinpacni} have shown how MRI derived pace measures can capture functional change. Siamese architectures are particularly suited to this setting because they model this change directly \citep{kim2025pace2}.

Several methodological challenges remain. Existing longitudinal approaches are limited by prediction accuracy, short interval ranges, and validation against established markers of neurodegenerative disease \citep{wegmann2025pace3, yin2025pace1, kim2025pace2}. Furthermore, \citet{peng2021accurate} showed how Soft Label Distribution (SLD) learning can outperform regression approaches when modelling brain-PAD but distributional objectives have not been considered in longitudinal approaches. SLD allows age to be represented as an ordered target and supports distance-aware distribution losses \citep{gao2017deep, hou2016squared, niu2016ordinal, geng2016label}. Recent ordinal work from \citet{shah2024ordinal} has shown distance-aware regularisation can preserve age ordering within the learned representation. Distributional objectives further allow ordinal structure to be applied directly on the predicted probabilities, while supporting uncertainty in estimations. Additionally, SLD approaches for brain age tend to apply the Kullback-Leibler (KL). KL-divergence does not explicitly model distances between predicted temporal intervals and becomes unstable when predicted probabilities approach certainty. Cramér distance provides an alternative ordinal loss by comparing cumulative distributions, allowing prediction errors to be penalised according to their distance from the target. Calibration and uncertainty are particularly important if longitudinal predictions are to be interpreted biologically and clinically.

To address these methodological challenges, we propose Brain-Predicted Age Acceleration (Brain-PACE), a Siamese deep learning framework for directly estimating longitudinal Pace, $P$, of brain ageing from paired T1w MRI. Relative to cross-sectional brain-PAD methods, Brain-PACE is designed to directly predict pace, $P$, by comparing baseline and follow-up scans using difference embeddings. We incorporate spatial attention to weight informative spatial embeddings and predict distributional targets rather than a single scalar value (applying Cram\'er distance as a robust ordinal loss function). This probabilistic approach allows predictive uncertainty to be estimated alongside longitudinal change. We evaluate Brain-PACE across hold out and independent healthy datasets, comparing its performance with standard and adapted cross-sectional brain-PAD approaches. We first assess whether Cram\'er distance and spatial attention improve longitudinal interval prediction, calibration, and generalisation. We then test whether directly modelling change provides additional benefit over estimating pace indirectly from two cross-sectional brain age predictions. Finally, we evaluate whether the resulting pace measure captures biologically meaningful information by examining its relationships with functional decline and regional tau burden in individuals with Mild Cognitive Impairment (MCI).

% \FloatBarrier 
\section{Methods}\label{}
\subsection{Datasets and Preprocessing}
Demographic, cognitive and imaging data used in the preparation of this article were obtained from the Alzheimer's Disease Neuroimaging Initiative (ADNI) database \citep{adni}. The ADNI was launched in 2003 and the original goal of ADNI was to test whether serial MRI, positron emission tomography (PET), other biological markers, and clinical and neuropsychological assessment can be combined to measure the progression of MCI and early AD. Data were also provided in part by the Open Access Series of Imaging Studies (OASIS-3) \citep{oasis}, and data were collected from the Australian Imaging, Biomarkers and Lifestyle (AIBL) study group. AIBL study methodology has been reported previously \citep{aibl}. Imaging and cognitive data for ADNI, OASIS-3, and AIBL were downloaded in January 2026. Independent healthy evaluation cohorts were also obtained from the Wisconsin Registry for Alzheimer's Prevention (WRAP) \citep{johnson2018wisconsin}, and Mayo Clinic Study of Ageing (MCSA) \citep{mcsa}. 

\begin{table}[!htbp]
\centering
\caption{Summary of subject demographics and scan intervals across all datasets. Ages and follow-up interval ($\Delta T$) are reported in years, with mean and standard deviation.}
\label{tab:demographics}
\begin{adjustbox}{max width=\textwidth}
\begin{tabular}{lllcccc}
\toprule
Group & Dataset & Baseline Age & Age Range & Mean $\Delta T$ & $n$ Subjects & $n$ Pairs \\
\midrule
Healthy Train+Validation
& ADNI, AIBL, OASIS
& 71.5 (6.6)
& 56.3--91.3
& 3.1 (1.7)
& 711
& 2780 \\

Healthy Test
& ADNI, AIBL, OASIS
& 72.6 (7.5)
& 56.1--91.4
& 3.2 (1.8)
& 81
& 363 \\

External Healthy
& MCSA
& 66.9 (6.9)
& 56.1--84.8
& 3.1 (1.6)
& 52
& 87 \\

External Healthy
& WRAP
& 63.6 (4.8)
& 57.0--78.0
& 3.6 (1.8)
& 84
& 316 \\

Evaluation MCI
& ADNI
& 74.8 (7.4)
& 57.7--90.0
& 2.3 (1.4)
& 68
& 121 \\
\bottomrule
\end{tabular}
\end{adjustbox}
\end{table}

Inline with control group criteria from \citet{petersen2010alzheimer}, healthy participants (for training and test sets) required Clinical Dementia Rating of $0$, Mini-Mental State Examination scores greater than $24$, and the absence of confounding health conditions. To maximise sample size, we applied dense longitudinal pairing, extracting all unique (chronological) scan combinations per subject with an interval $\Delta T$  between $0.5$ and $7.5$ years. Groups were defined at the subject level to prevent data leakage, providing a maximum sample of $3143$ image pairs (ADNI; $2635$ AIBL; $285$ OASIS; $223$) from $792$ unique subjects (ADNI; $530$ AIBL; $110$ OASIS; $152$). For external validation of healthy participants, the MCSA and WRAP were additionally included, alongside a separate ADNI sample of participants MCI. MCI participant data was used to assess model sensitivity to biomarkers of neurodegeneration and pathological ageing trajectories. WRAP and MCSA eligibility required a CDR that never exceeded $0$ at any available visit and exclusion of participants with recorded current or past conditions (including but not limited to cerebrovascular disease, major neurological or neurodegenerative diseases, brain injury, or major psychiatric disorders). Full demographics are summarised in Table \ref{tab:demographics}.

\begin{figure}[!htbp]
\centering
\includegraphics[width=\textwidth]{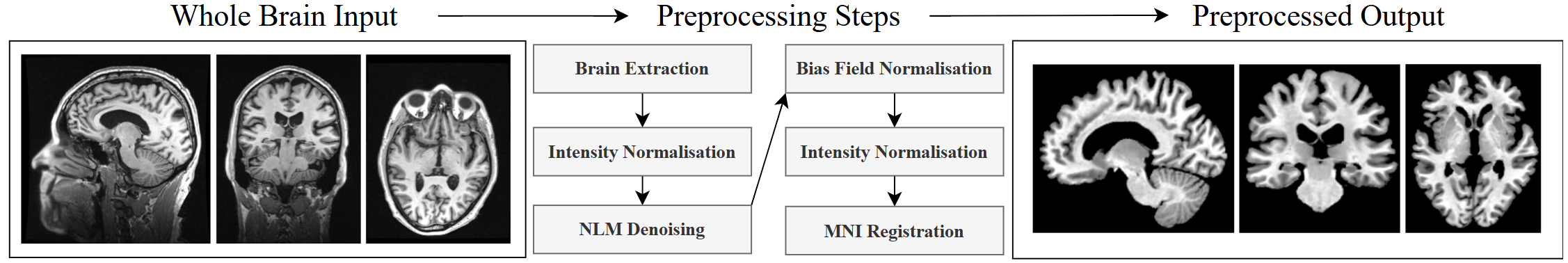}
\caption{Overview of the MRI preprocessing steps illustrating the transformation of raw T1w inputs (left) to fully registered and normalised outputs (right). Key Steps involve brain extraction, non-local means denoising, bias field correction, and spatial registration to standardised Montreal Neurological Institute space.}
\label{fig:preproc}
\end{figure}

The applied T1w preprocessing pipeline is represented in Figure \ref{fig:preproc}. To normalise input images and mitigate site-specific noise, structural MRI scans were preprocessed by following the multistep framework from \citet{rajabli2025brain}. Brain extraction was initially performed using SynthStrip \citep{hoopes2022synthstrip}, followed by intensity normalisation, non-local means denoising \citep{manjon2010adaptive}, and N4 bias field correction \citep{tustison2010n4itk}. Following a second iteration of intensity normalisation, volumes were affine-registered to a 1mm standard Montreal Neurological Institute (MNI152) template using EasyReg \citep{easyreg}. Non-brain tissue was removed using grey and white matter masks from SynthSeg, and volumes were cropped to $167\times212\times160$ voxels \citep{billot2023robust}. 

\FloatBarrier 
\subsection{Network Architecture}
\begin{figure}[!htbp]
\centering
\includegraphics[width=\textwidth]{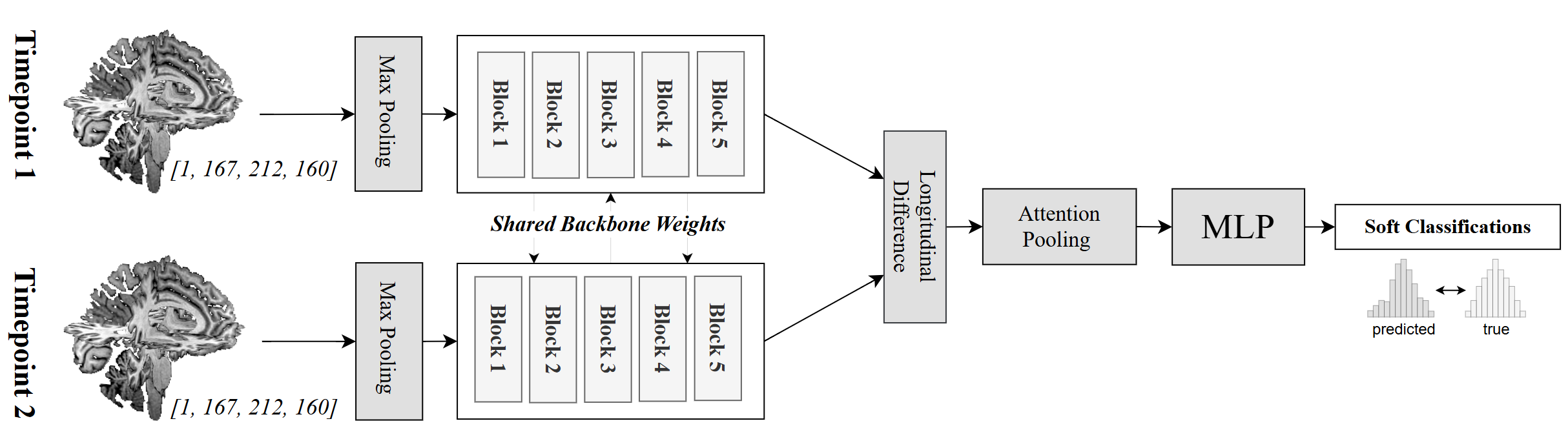}
\caption{Proposed Siamese deep learning architecture for longitudinal brain age estimation. Longitudinal T1-weighted image pairs (Timepoint $1$ and Timepoint $2$) are processed through a shared SFCN-style backbone. Resulting timepoint-specific feature vectors are fused via subtraction. The difference vector is then passed to the spatial attention pooling module and Multi-Layer Perceptron head to finally produce a soft classification distribution.} \label{fig:architecture}
\end{figure}

 Following the LILAC approach, our longitudinal framework uses a Siamese deep learning architecture to extract temporal embeddings directly from baseline and follow-up image pairs. The Simple Fully Convolutional Network (SFCN) backbone was implemented because of its reported ability to effectively capture brain MRI features \citep{peng2021accurate}. An initial 3D Max Pooling layer (kernel size and stride of 2) was first applied to reduce input dimensionality. The backbone comprises five convolutional blocks, each containing a $3\times3\times3$ 3D convolution, 3D Batch normalisation, and a ReLU activation function. Standard SFCN architectures apply a Pooling layer after the convolution layers, however we opted to remove this in favour of downstream spatial attention. Once the Siamese network generates embeddings from baseline and follow-up images, they are directly fused via subtraction. We then apply the spatial attention pooling directly onto the difference vector. This involves a $1\times1\times1$ convolution to the final feature map to generate spatial logits, which are normalised via Softmax to compute the final timepoint-specific feature vector. Following recent LILAC+ methods, learnt embeddings are passed into a MLP head which has two hidden layers containing $2048$ and $512$ units. The proposed architecture is illustrated in Figure ~\ref{fig:architecture}. 

\begin{figure}[!htbp]
\centering
\includegraphics[width=\textwidth]{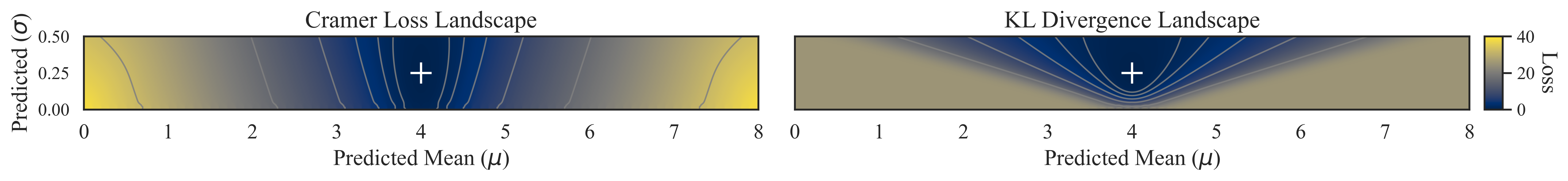}
\caption{Comparison of the loss landscapes for the Cram\'er distance and
KL divergence. Each heat map shows the loss as a function of the predicted
mean $\mu$ and standard deviation $\sigma$ relative to a fixed target
distribution, marked by a cross at $\mu=4.0$ and $\sigma=0.25$.
The Cram\'er distance landscape (\textbf{left}) exhibits a smooth,
valley-like topology, whereas the KL-divergence landscape
(\textbf{right}) is more sharply constrained. Contour levels are spaced
at powers of two, with higher levels indicating greater loss.}
\label{fig:loss_landscape}
\end{figure}

Rather than predicting a single continuous scalar, the temporal gap is mapped across 80 discrete bins with a resolution of 0.1 years per bin, allowing the target interval range of [$0, 8$] years. In prior work, KL-divergence is applied even though it does not model distances between discrete age bins and therefore lacks the directional information that is useful for ordinal age modelling. It also shows instability when predicted probabilities trend towards certainty. To address these limitations, we apply a  Cram\'er distance loss function which minimises the squared difference between the predicted and target Cumulative Distribution Functions (CDFs), where Loss ($L$): 
\begin{equation}
L = \sum_{j} \left( \mathrm{CDF}_p(j) - \mathrm{CDF}_q(j) \right)^2,
\quad \text{where} \quad
\mathrm{CDF}_p(j) = \sum_{k \le j} p_k
\end{equation}
% \newpage
and $p_j$ and $q_j$ represent the predicted and target discrete probability masses at bin $j$, respectively. To visualise the optimisation advantage of this approach, Figure ~\ref{fig:loss_landscape} visualises the loss landscapes for KL-divergence and the proposed Cram\'er distance metric, using an example target prediction mean ($\mu$) of $4$ and distribution spread ($\sigma$) of $0.25$. While KL-divergence penalises closer misses to the same level as further misses,  Cram\'er distance instead forms a smoother, constrained loss landscape. This topographical stability guides the network toward the mean and enables flexible prediction of distribution spread. 

 % Figure 1: Siamese Architecture

To reduce overfitting, final predictions were generated using a cross-validation ensemble approach, where outputs of the models across 5 folds were averaged to calculate the final pace estimation for each longitudinal pair. Models were implemented in PyTorch and trained for 50 epochs on an NVIDIA RTX 6000 GPU before being stored for the downstream ensemble framework. A cosine annealing learning rate scheduler with a $5\%$ linear warmup phase (maximum learning rate of 0.05), and SGD optimiser and a batch size of 16 was used. 

For model comparisons, repeated scan-pair results were simplified to a single score using the participant median in order for each participant to receive equal weight. Paired model differences were evaluated with 10,000 bootstrap resamples, with bootstrap $p$ values and $95\%$ confidence intervals. False discovery rate (FDR) correction was applied across pairwise model comparisons separately within each dataset, and results below use the resulting FDR $q$ values. Pearson $r$ and $R^2$ were calculated after participant level median aggregation. For the architecture comparisons, comparisons were performed on the raw predictions to be consistent with previously reported results. To maintain consistency with previously reported results, performance against similar methods is derived at the pair level.

% \newpage 
\subsection{Model Performance and Uncertainty Evaluation}
To define the performance metrics used in the evaluation, $y_i$ represents the observed chronological interval and $\hat{y}_i$ represents the predicted interval for a given longitudinal pair $i$, over a total of $N$ pairs. Model performance was evaluated using the Continuous Ranked Probability Score (CRPS) \citep{gneiting2007probabilistic}. CRPS compares the complete predicted cumulative distribution with the observed scan interval. Using numerical integration over the bin width $\Delta x$, the discrete implementation is calculated across the total number of bins, $N_{bins}$ where $j$ defines the specific bin index:
\begin{equation}
CRPS = \sum_{j=1}^{N_{bins}} (CDF_{predicted}(j) - CDF_{observed}(j))^2 \Delta x
\end{equation}
Scores are reported with lower CRPS values indicating predictive distributions that are both more accurate and more concentrated around the observed interval. CRPS was selected because it evaluates the full probabilistic prediction rather than only its expected value. Performance was additionally assessed using mean absolute error (MAE), mean squared error (MSE), absolute percentage error (APE), signed error, and the pace ratio ($P$). MAE and MSE quantify absolute and squared differences, respectively, between the predicted interval $\hat{y}$ and observed interval $y$, while APE expresses the absolute error relative to the observed interval. Signed error, defined as $\hat{y}-y$, was used to assess over/under prediction. Pace was defined as $P=\hat{y}/y$, where $P=1$ indicates agreement with chronological ageing.

Predictive uncertainty was quantified using the Shannon entropy of each predicted probability distribution. Raw entropy ranged from $0$, indicating complete certainty, to a theoretical maximum of $\ln(80)=4.38$, corresponding to a uniform distribution across all 80 bins. Entropy was normalised by this maximum and expressed as a percentage, where $0\%$ represented complete certainty and $100\%$ represented maximum uncertainty. Empirical coverage was evaluated for the $50\%$, $80\%$, $90\%$, and $95\%$ central prediction intervals by calculating the proportion of observed intervals contained within each interval. 

\subsection{Integrated Gradient Interpretability Analysis}
Voxel-wise attribution maps were generated using Integrated Gradients (IG) implemented from the Python Captum library. For each longitudinal pair, the baseline (earlier time point) scan was held fixed in the first Siamese backbone, while the follow-up scan was interpolated from the baseline reference. This therefore details which voxel level changes from time point one to time point two contribute to the predicted temporal interval. The final attribution score for each voxel was calculated by scaling the model's sensitivity by the actual change that occurred at that location:
\begin{equation}
    \text{Voxel Attribution} = (\text{Follow-up} - \text{Baseline Intensity}) \times \text{Average Model Sensitivity}
    \label{eq:integrated_gradients}
\end{equation}
Here, the average model sensitivity represents the accumulated gradients computed across the entire interpolated path between the two scans. This ensures that regions with no morphological change receive zero attribution, precisely isolating the structural alterations that drive the network's temporal prediction. From testing, we found that the IG model converged at 30 steps, and this was therefore applied across each fold. Attributions were computed across all five-folds and averaged across models. To obtain a group attribution map, each pair-specific map was normalised by the sum of its absolute voxel attribution values to ensure that individual scans contributed equally regardless of overall attribution magnitude. Normalised maps were then averaged across all valid longitudinal pairs. The absolute value of the resulting average map was calculated and spatially smoothed using a Gaussian kernel with a full width at half maximum of $2\mathrm{mm}$. For anatomical interpretation, contiguous positive-attribution clusters exceeding a voxel threshold of $1\times10^{-6}$ were analysed. 

\subsection{Aligned Brain Age Model}
To enable direct comparison between cross-sectional brain-PAD estimation and longitudinal Brain-PACE, we trained a brain age model using the same backbone architecture with spatial attention pooling, soft-classifications and Cram\'er distance loss. In line with the longitudinal methods, predictions from the five cross validation models were averaged to obtain ensemble brain age estimates. Prediction bias was corrected using linear regression fitted to the combined validation predictions, with brain-PAD as the error target (predicted -- chronological age). The fitted correction was then applied unchanged to test set predictions. Longitudinal change was estimated as the difference between the brain age predictions at follow-up and baseline. Consistent reporting and bootstrap procedures were applied to the indirectly derived pace estimates as mentioned in Section \textbf{2.2} for consistent reporting.

\subsection{Statistical Analysis}
To evaluate relationships between $P$ and established AD biomarkers, we performed a targeted statistical analysis on the MCI group. Raw pace was calculated as the ratio of the predicted temporal interval to the actual chronological interval between scans. To reduce dependence on scan interval, a linear healthy-reference model was fit to the temporal gap and signed prediction error in the internal healthy cohort using 5-fold group cross validation. Out-of-fold expected errors were then used for internal participants, this final model was applied unchanged to MCSA, WRAP, and ADNI. Healthy-adjusted pace was defined as $1+(\mathrm{observed~error}-\mathrm{expected~error})/\Delta T$. Repeated pair-level values were collapsed to a single median value per participant. Accelerated and decelerated ageing were defined using the 95th and 5th percentiles, respectively, of the hold out participant distribution. For comparison, the bias-corrected brain age model was evaluated as both an indirect pace measure and standard cross-sectional brain-PAD.

We computed partial Pearson correlations ($r$) between participant-level pace and specific baseline features \citep{adni}, controlling direct and indirect pace associations for baseline age and median temporal gap. Brain-PAD associations were controlled for baseline age. Clinical and cognitive assessments included the Clinical Dementia Rating Sum of Boxes (CDR-SB), Mini-Mental State Examination (MMSE), Functional Assessment Questionnaire (FAQ), Geriatric Depression Scale (GDS), and the Alzheimer's Disease Assessment Scale-Cognitive Subscale 13 (ADAS-Cog 13). Structural volumetric measures and tau standard uptake value ratios (SUVR) were analysed in a select set of anatomical brain regions with known links to AD pathology: the entorhinal cortex, brainstem, posterior cingulate cortex, precuneus, hippocampus, and thalamus proper. FDR correction was applied across the tested features within each method, with statistical significance defined at $q<0.05$.

% \FloatBarrier  
\section{Results}\label{}
\subsection{Brain-PACE Performance Evaluation and Loss Comparison}

We evaluated four Brain-PACE configurations combining Cram\'er distance or KL-divergence loss with either spatial attention or global average pooling. Probabilistic performance was assessed using CRPS, with lower values indicating a predicted distribution that was both more accurate and more concentrated around the observed scan interval. Performance was additionally evaluated with MAE, APE, signed error, and pace $P$ (defined as the predicted interval divided by the observed chronological interval). A $P$ of $1$ indicates agreement with the observed rate of change. Full results are reported in Table \ref{tab:performance_1}.

\begin{table}[!htbp]
\centering
\caption{Performance across the various architecture ablations in the primary internal, Mayo Clinic Study of Ageing (MCSA), Wisconsin Registry for Alzheimer's Prevention (WRAP) test sets. Lower Continuous Ranked Probability Score (CRPS), Mean Absolute Error (MAE), and Absolute Percentage Error (APE) values indicate better performance. Signed error values closer to $0$ indicate lower bias, with positive values indicating over prediction and negative values indicating under prediction. Pace $P$ ratios closer to $1$ indicate better agreement with the observed rate of change. The best result within each group is shown in bold.}
\label{tab:architecture_ablation}
\begin{adjustbox}{max width=\textwidth}
\begin{tabular}{lllccccc}
\toprule
Group &  Dataset&Model & CRPS & MAE & APE & Signed error & Pace ratio \\
\midrule
\multirow{4}{*}{Internal Healthy}
&  \multirow{4}{*}{ADNI,OASIS,AIBL}&CR + Attn & \textbf{0.471} & \textbf{0.663} & \textbf{0.272} & \textbf{0.017} & \textbf{1.110} \\
&  &CR + Avg & 0.525 & 0.734 & 0.316 & 0.268 & 1.196 \\
&  &KL + Attn & 0.508 & 0.705 & 0.319 & 0.240 & 1.208 \\
&  &KL + Avg & 0.649 & 0.922 & 0.462 & 0.672 & 1.405 \\
\midrule
\multirow{4}{*}{External Healthy}&  \multirow{4}{*}{MCSA}&CR + Attn & \textbf{0.441} & \textbf{0.615} & \textbf{0.240} & \textbf{-0.012} & \textbf{1.074} \\
&  &CR + Avg & 0.529 & 0.744 & 0.313 & 0.449 & 1.228 \\
&  &KL + Attn & 0.455 & 0.638 & 0.266 & 0.302 & 1.179 \\
&  &KL + Avg & 0.744 & 1.055 & 0.468 & 0.946 & 1.441 \\
\midrule
\multirow{4}{*}{External Healthy}&  \multirow{4}{*}{WRAP}&CR + Attn & \textbf{0.803} & \textbf{1.066} & \textbf{0.317} & -0.402 & \textbf{0.995} \\
&  &CR + Avg & 0.877 & 1.129 & 0.371 & \textbf{-0.054} & 1.089 \\
&  &KL + Attn & 0.885 & 1.162 & 0.345 & -0.396 & 1.015 \\
&  &KL + Avg & 0.879 & 1.202 & 0.425 & 0.177 & 1.210 \\
\bottomrule
\end{tabular}
\end{adjustbox}
\label{tab:performance_1}
\end{table}

The Cram\'er-attention model numerically achieved the lowest CRPS in both internal and external test set groups when combined with spatial attention. In the internal group, Cram\'er-attention achieved a CRPS of $0.471$ years compared with $0.508$ years for KL-attention ($q=0.012$). The corresponding MAE was numerically lower in the Cram\'er-attention model ($0.663$ versus $0.705$ years), but was not significant after FDR correction ($q=0.118$). Significant improvements were observed for APE ($0.272$ versus $0.319$, $q=0.002$), signed error ($0.017$ versus $0.240$ years, $q<0.001$), and $P$ ($1.110$ versus $1.208$, $q<0.001$). The benefits of spatial attention were further evaluated within the Cram\'er models. Attention numerically reduced CRPS from $0.525$ to $0.471$ years ($q=0.058$) and the MAE reduction from $0.734$ to $0.663$ years was also not significant ($q=0.093$). APE improved from $0.316$ to $0.272$ ($q=0.026$), signed error was reduced from $0.268$ to $0.017$ years ($q<0.001$), and $P$ moved closer to 1, from $1.196$ to $1.110$ ($q<0.001$). These findings indicate that the combination of Cram\'er loss and spatial attention improved probabilistic performance and reduced prediction bias, although not every error metric remained significant after correction.

In the MCSA external test set, Cram\'er-attention again produced numerically stronger (though non-significant) improvements over KL-attention for CRPS ($0.441$ versus $0.455$ years; $q=0.590$), MAE ($0.615$ versus $0.638$ years; $q=0.481$), and APE ($0.240$ versus $0.266$; $q=0.241$). However, Cram\'er-attention produced significantly lower signed error ($-0.012$ versus $0.302$ years; $q<0.001$) and a $P$ closer to 1 ($1.074$ versus $1.179$; $q<0.001$). Additionally, replacing global average pooling with attention numerically reduced CRPS from $0.529$ to $0.441$ years ($q=0.136$) and MAE from $0.744$ to $0.615$ years ($q=0.136$). APE decreased from $0.313$ to $0.240$ ($q=0.018$), signed error from $0.449$ to $-0.012$ years ($q<0.001$), and $P$ from $1.228$ to $1.074$ ($q<0.001$). These results indicate that the Cram\'er approach benefited from spatial attention particularly for percentage error, bias, and pace ratio, and generalised well from the internal group. WRAP provided a second independent healthy test of generalisation. Cram\'er-attention achieved lower CRPS ($0.803$ versus $0.885$ years; $q<0.001$), MAE ($1.066$ versus $1.162$ years; $q=0.007$), MSE ($2.089$ versus $2.459$ years$^2$; $q=0.002$), and APE ($0.317$ versus $0.345$; $q=0.014$) than KL-attention, while its pace ratio was closer to 1 ($0.995$ versus $1.015$; $q=0.014$). Signed error was similar ($-0.402$ versus $-0.396$ years; $q=0.173$). Within the Cram\'er models, spatial attention significantly improved CRPS, MAE, MSE, APE, signed error, and pace ratio after FDR correction (all $q<0.05$).

The MCI group was interpreted separately as positive signed error and $P$ above $1$ may reflect accelerated or abnormal brain ageing rather than just prediction error. The Cram\'er-attention model showed lower values across all reported metrics. Compared with KL-attention, it achieved lower CRPS ($0.571$ versus $0.695$ years), MAE ($0.809$ versus $0.957$ years), APE ($0.520$ versus $0.628$), signed error ($0.497$ versus $0.746$ years), and $P$ closer to 1 ($1.434$ versus $1.571$); all comparisons remained significant after FDR correction ($q<0.001$). Spatial attention also lowered the Cram\'er model error relative to global average pooling across CRPS, MAE, absolute percentage error, signed error, and $P$ (all $q<0.001$). 

Overall, the Cram\'er metric loss consistently performed well relative to KL-divergence, particularly when combined with spatial attention. Its strongest advantages were observed for probabilistic accuracy and pace calibration. While some differences did not remain statistically significant in the internal and MCSA cohorts after FDR correction, the independent WRAP cohort showed significant improvements in CRPS, MAE, MSE, APE, and pace ratio for Cram\'er-attention over KL-attention. When evaluated at the pair level (to remain directly comparable with previously reported methods that do not use subject-level aggregation) Brain-PACE showed the strongest overall error performance compared to other comparable methods. As shown in Table \ref{tab:performance_2}, MAE decreased from $0.99$ years for the reported LILAC+ model to $0.66$ years for Brain-PACE, while MSE decreased from $1.97$ to $0.76$ years$^2$. When implemented locally using the same test dataset and preprocessing, Brain-PACE still showed improvements of 14.9\% decrease in MAE and 28.6\% decrease in MSE. Brain-PACE further showed improved model fit $r=0.88$ (5.6\% improvement) and $R^2=0.77$ (13.5\% improvement). Our findings therefore support Cram\'er loss as the more robust objective, while acknowledging that the significance of the advantage varied across cohorts in some cases.

\begin{table}[!htbp]
\caption{Comparison of the published LILAC and LILAC+ results reported by \citet{wegmann2025pace3} with local implementations LILAC/LILAC+ models and Brain-PACE. LILAC/LILAC+ results were obtained using the datasets and preprocessing described by \citet{wegmann2025pace3}, whereas the local LILAC/LILAC+ and Brain-PACE models were evaluated using datasets and preprocessing described in this work. To maintain consistency with previously reported results, performance here is derived  from the pair level results. Metrics of Mean Absolute Error (MAE), Mean Squared Error (MSE), Pearson correlation ($r$), and variance explained ($R^2$) are included.}
\label{tab:performance_2}
\centering
\begin{adjustbox}{max width=\textwidth}
\begin{tabular}{llcccccc}
\toprule
Model & Approach & Baseline age & Mean $\Delta T$
& MAE & MSE & $r$ & $R^2$ \\
\midrule

\multicolumn{8}{l}{\textit{Previously reported results \citet{wegmann2025pace3}}} \\

LILAC
& Linear head
& 68.7 (9.8) & 4.18 (--)
& 1.16 & 2.29 & 0.86 & 0.67 \\

LILAC+
& MLP head
& 68.7 (9.8) & 4.18 (--)
& 0.99 & 1.97 & 0.86 & 0.71 \\

\midrule

\multicolumn{8}{l}{\textit{Local implementations, datasets and image preprocessing}} \\

LILAC (Local)
& Linear head
& 72.6 (7.5) & 3.24 (1.8)
& 0.899 & 1.390 & 0.837 & 0.581 \\

LILAC+ (Local)
& MLP head
& 72.6 (7.5) & 3.24 (1.8)
& 0.779 & 1.059 & 0.832 & 0.680 \\

\midrule

Brain-PACE (Ours)
& MLP + KL + Attn
& 72.6 (7.5) & 3.24 (1.8)
& 0.705
& 0.844
& 0.874
& 0.745 \\

Brain-PACE (Ours)
& MLP + Cram\'er + Attn
& 72.6 (7.5) & 3.24 (1.8)
& \textbf{0.663}
& \textbf{0.756}
& \textbf{0.879}
& \textbf{0.772} \\

\bottomrule
\end{tabular}
\end{adjustbox}
\end{table}

% \FloatBarrier 
\subsection{Uncertainty Evaluation}
The Cram\'er-attention model produced significantly lower normalised entropy than the KL-attention model in all groups (mean differences of $-9.85$, $-10.21$, $-10.77$, and $-9.49$ percentage points for Internal, MCSA, WRAP, and ADNI, respectively; all FDR-adjusted $q<0.001$). Coverage was closest to the nominal intervals in the internal and MCSA groups. In WRAP, Cram\'er-attention also showed higher empirical coverage than KL-attention at all four intervals, with the 90\% interval difference remaining significant after FDR correction ($82.7\%$ versus $73.8\%$, $q=0.044$). Coverage decreased most clearly in ADNI, indicating that both models underestimated uncertainty under pathological ageing; Cram\'er-attention showed higher coverage than KL-attention at every interval, with significant paired differences across all four ADNI coverage levels. See Table \ref{tab:coverage} for the full uncertainty and coverage results.

\begin{table}[!htbp]
\centering
\caption{Participant-weighted uncertainty and empirical coverage across the internal healthy, Mayo Clinic Study of Ageing (MCSA), Wisconsin Registry for Alzheimer's Prevention (WRAP), and Mild Cognitive Impairment (MCI) test sets for Cram\'er (CR) and Kullback--Leibler (KL) spatial-attention models. Entropy is reported as the mean normalised entropy with 95\% bootstrap confidence intervals. Coverage values represent participant-weighted empirical coverage for the 95\%, 90\%, 80\%, and 50\% central prediction intervals.}
\label{tab:coverage}

\begin{adjustbox}{max width=\textwidth}
\begin{tabular}{lllccccc}
\toprule
Group & Dataset & Model & Entropy (\%) & 95\% & 90\% & 80\% & 50\% \\
\midrule

\multirow{2}{*}{Internal Healthy}
& \multirow{2}{*}{ADNI, OASIS, AIBL}
& CR\_attn & 62.9 [61.9, 63.8] & 95.7 & 94.4 & 83.3 & 51.9 \\
&
& KL\_attn & 72.7 [71.4, 74.1] & 89.5 & 87.0 & 77.2 & 46.9 \\

\midrule

\multirow{2}{*}{External Healthy}
& \multirow{2}{*}{MCSA}
& CR\_attn & 62.4 [61.1, 63.6] & 91.3 & 87.5 & 76.0 & 43.3 \\
&
& KL\_attn & 72.6 [71.0, 74.2] & 86.5 & 79.8 & 74.0 & 37.5 \\

\midrule

\multirow{2}{*}{External Healthy}
& \multirow{2}{*}{WRAP}
& CR\_attn & 64.7 [63.6, 65.8] & 86.9 & 82.7 & 74.4 & 42.3 \\
&
& KL\_attn & 75.5 [73.9, 77.0] & 83.9 & 73.8 & 65.5 & 38.1 \\

\midrule

\multirow{2}{*}{Evaluation MCI}
& \multirow{2}{*}{ADNI}
& CR\_attn & 62.7 [61.5, 63.9] & 77.9 & 72.1 & 61.0 & 36.0 \\
&
& KL\_attn & 72.2 [70.7, 73.6] & 62.5 & 55.9 & 45.6 & 23.5 \\

\bottomrule
\end{tabular}
\end{adjustbox}
\end{table}

% \FloatBarrier 
\subsection{Integrated Gradient Explainability}
\begin{figure}[!htbp]
\centering
\includegraphics[width=\textwidth]{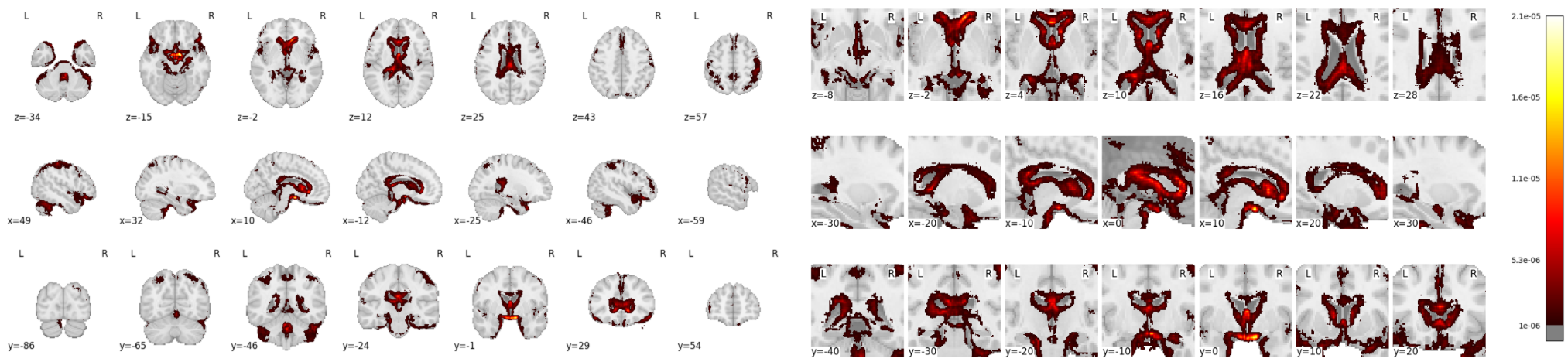}
\caption{Left: group-level whole-brain template absolute IG attribution map shown across representative axial, sagittal, and coronal slices. Right: focused ventricular absolute IG attribution map. Warmer colours indicate regions with greater influence on the model prediction.}
\label{fig:IG2}
\end{figure}

The group-level attribution map presented in Figure \ref{fig:IG2} shows a distributed pattern across anatomically specific cortical and subcortical regions. The strongest contributions were observed in the frontal and temporal poles, lateral occipital cortex, precentral and postcentral gyri, and brainstem. Additional attribution involved the orbitofrontal, superior and middle frontal, precuneus, superior parietal, anterior cingulate, and paracingulate regions. Subcortical and medial temporal contributions included the bilateral thalami, right caudate, anterior parahippocampal gyrus, subcallosal cortex, and posterior cingulate cortex. Further attribution was observed in the lingual, supramarginal, angular, inferior temporal, and occipital regions. Overall, the predicted longitudinal intervals were informed by distributed changes across frontal, temporal, parietal, occipital, sensorimotor, subcortical, and brainstem structures rather than a single anatomical region.

% \FloatBarrier  
\subsection{Brain Age Model Performance}
The cross-sectional brain age model achieved an MAE of $2.64$ years internally, $2.21$ years in MCSA, $2.22$ years in WRAP, and $3.06$ years in the ADNI MCI group. We then compared longitudinal pace derived indirectly from brain age predictions with those estimated directly by Brain-PACE. In the internal group, Brain-PACE showed numerically lower MAE ($0.663\pm0.564$ versus $0.806\pm0.750$ years; $q=0.153$) and MSE ($0.756\pm1.509$ versus $1.209\pm2.505$ years$^2$; $q=0.153$), although neither difference remained significant after FDR correction. Signed error was also lower for Brain-PACE ($0.017\pm0.871$ versus $0.307\pm1.057$ years; $q=0.083$), while mean pace $P$ was similar between Brain-PACE and the indirect brain age approach ($1.110\pm0.366$ versus $1.089\pm0.384$; $q=0.817$).

In MCSA, the two approaches performed similarly for MAE ($0.615\pm0.499$ versus $0.642\pm0.519$ years; $q=0.847$), MSE ($0.625\pm1.177$ versus $0.679\pm0.965$ years$^2$; $q=0.847$), and signed error ($-0.012\pm0.795$ versus $-0.011\pm0.829$ years; $q=0.771$). The indirect approach produced a mean $P$ numerically closer to 1 ($0.982\pm0.351$ versus $1.074\pm0.325$), although the difference was not significant after FDR correction ($q=0.771$). WRAP showed a similar pattern for conventional error measures. Brain-PACE had an MAE of $1.066\pm0.978$ years compared with $0.992\pm0.837$ years for the indirect approach ($q=0.134$), and an MSE of $2.089\pm3.715$ versus $1.682\pm2.571$ years$^2$ ($q=0.134$). Signed error differed significantly between approaches, with Brain-PACE showing greater underprediction ($-0.402\pm1.391$ versus $-0.029\pm1.299$ years; $q=0.021$). Mean pace was similar between Brain-PACE and the indirect approach ($0.995\pm0.462$ versus $1.004\pm0.463$; $q=0.355$).

In the MCI group, the indirect brain age approach showed numerically lower conventional error values than Brain-PACE, including MAE ($0.766\pm0.665$ versus $0.809\pm0.668$ years; $q=0.081$) and MSE ($1.025\pm1.711$ versus $1.097\pm1.950$ years$^2$; $q=0.081$). Signed error was significantly higher for Brain-PACE ($0.497\pm0.926$ versus $0.419\pm0.926$ years; $q=0.030$), and Brain-PACE also produced a significantly higher $P$ ($1.434\pm0.638$ versus $1.220\pm0.506$; $q<0.001$). Larger Brain-PACE deviations in the MCI group may reflect sensitivity to accelerated or disease-related brain ageing rather than poorer prediction accuracy alone. We evaluate this in the following section.

% \FloatBarrier  
\subsection{Healthy and Pathological Ageing}
Minimising scan-interval prediction error does not necessarily establish the utility of Brain-PACE. We therefore tested whether the resulting pace measure was associated with established markers of dementia across cognition, functional impairment, brain volume, and regional tau measures (including the posterior cingulate, precuneus, entorhinal cortex, hippocampus, thalamus, and brainstem).

We first evaluated the normalised pace score, $P$, across groups after accounting for the relationship between temporal gap and signed prediction error in the internal healthy reference cohort. For direct Brain-PACE, internal healthy participants showed a mean adjusted pace of $0.98$ ($\pm0.25$), with $4.9\%$ classified as accelerated and $4.9\%$ as decelerated by the empirical healthy thresholds. MCSA showed a similar profile, with a mean adjusted pace of $1.00$ ($\pm0.28$), $9.6\%$ accelerated and $5.8\%$ decelerated. WRAP showed a lower mean adjusted pace of $0.90$ ($\pm0.25$), with $4.8\%$ accelerated and $19.0\%$ decelerated. In contrast, the ADNI MCI group showed a higher mean adjusted pace of $1.33$ ($\pm0.62$), with $42.6\%$ classified as accelerated and $7.4\%$ as decelerated.

The indirectly derived brain age pace showed a similar mean adjusted pace in the internal healthy ($0.96\pm0.37$) and WRAP ($0.90\pm0.39$) groups, with MCSA showing a mean adjusted pace of $0.94$ ($\pm0.32$). In the ADNI MCI group, indirect pace was higher at $1.17$ ($\pm0.58$), with $20.6\%$ classified as accelerated and $8.8\%$ as decelerated. Thus, although both approaches showed a shift towards faster ageing in MCI, the proportion classified as accelerated was higher for direct Brain-PACE ($42.6\%$) than for indirectly derived brain age pace ($20.6\%$). The full results are provided in Table~\ref{tab:group_outliers}.

\begin{table}[!htbp]
\caption{Participant-level healthy-adjusted pace of brain ageing and proportions classified as decelerated or accelerated for direct Brain-PACE and indirectly derived brain age pace. Pace values are reported as mean (standard deviation) and median. Acceleration and deceleration are defined independently for each approach using the 95th and 5th percentiles of the corresponding internal healthy participant-level adjusted-pace distribution.}
\label{tab:group_outliers}
\centering
\begin{adjustbox}{max width=\textwidth}
\begin{tabular}{lcccccc}
\toprule
Group & Participants & Mean adjusted $P$ & Median adjusted $P$ & Decelerated & Accelerated \\
\midrule

\multicolumn{6}{l}{\textbf{Method: Brain-PACE Direct Pace}} \\

Internal Healthy
& 81 & 0.98 (0.25) & 0.96 & 4.9\% & 4.9\% \\

MCSA Healthy
& 52 & 1.00 (0.28) & 0.96 & 5.8\% & 9.6\% \\

WRAP Healthy
& 84 & 0.90 (0.25) & 0.91 & 19.0\% & 4.8\% \\

ADNI MCI
& 68 & 1.33 (0.62) & 1.16 & 7.4\% & 42.6\% \\

\midrule

\multicolumn{6}{l}{\textbf{Method: Brain-PAD Indirect Pace}} \\

Internal Healthy
& 81 & 0.96 (0.37) & 0.93 & 4.9\% & 4.9\% \\

MCSA Healthy
& 52 & 0.94 (0.32) & 0.92 & 7.7\% & 5.8\% \\

WRAP Healthy
& 84 & 0.90 (0.39) & 0.90 & 13.1\% & 6.0\% \\

ADNI MCI
& 68 & 1.17 (0.58) & 1.17 & 8.8\% & 20.6\% \\

\bottomrule
\end{tabular}
\end{adjustbox}
\end{table}

\begin{figure}[!htbp]
\centering
\includegraphics[width=\textwidth]{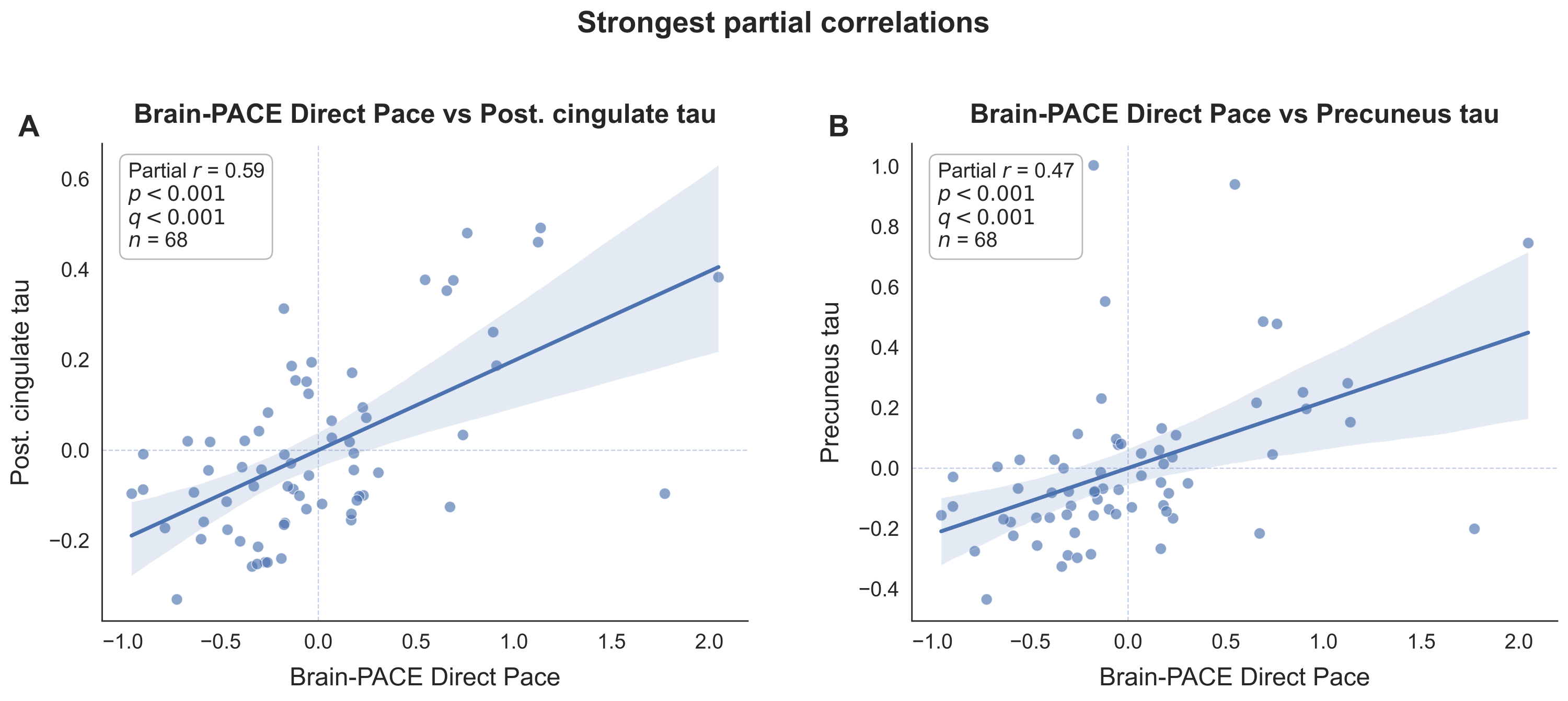}
\caption{Strongest partial correlations between direct Brain-PACE and regional tau burden in the ADNI Mild Cognitive Impairment group. (A) Direct Brain-PACE was positively associated with posterior cingulate tau SUVR (partial $r=0.59$, $q<0.001$). (B) Direct Brain-PACE was positively associated with precuneus tau SUVR (partial $r=0.47$, $q<0.001$). Partial correlations control for baseline age and median temporal gap ($n=68$). Shaded regions indicate the 95\% confidence interval.}
\label{fig:top_tau_correlations}
\end{figure}

Within the ADNI MCI group, participant brain-PACE was significantly associated with markers of tau pathology, cognition, and functional impairment after controlling for baseline age and median temporal gap. Signifiant associations were seen in the posterior cingulate ($r=0.59$, $q<0.001$), precuneus ($r=0.47$, $q<0.001$), and entorhinal cortex ($r=0.37$, $q=0.011$) tau SUVR. The two strongest associations, with posterior cingulate and precuneus tau burden, are shown in Figure~\ref{fig:top_tau_correlations}. Faster pace was also associated with greater functional impairment on the FAQ ($r=0.35$, $q=0.015$), poorer performance on the ADAS13 ($r=0.30$, $q=0.034$), and greater impairment on the CDR sum of boxes ($r=0.32$, $q=0.027$). The MMSE association was negative ($r=-0.26$) but did not survive FDR correction ($q=0.087$). No regional brain-volume measures remained significantly associated with pace after FDR correction.

\begin{figure}[!htbp]
\centering
\includegraphics[width=\textwidth]{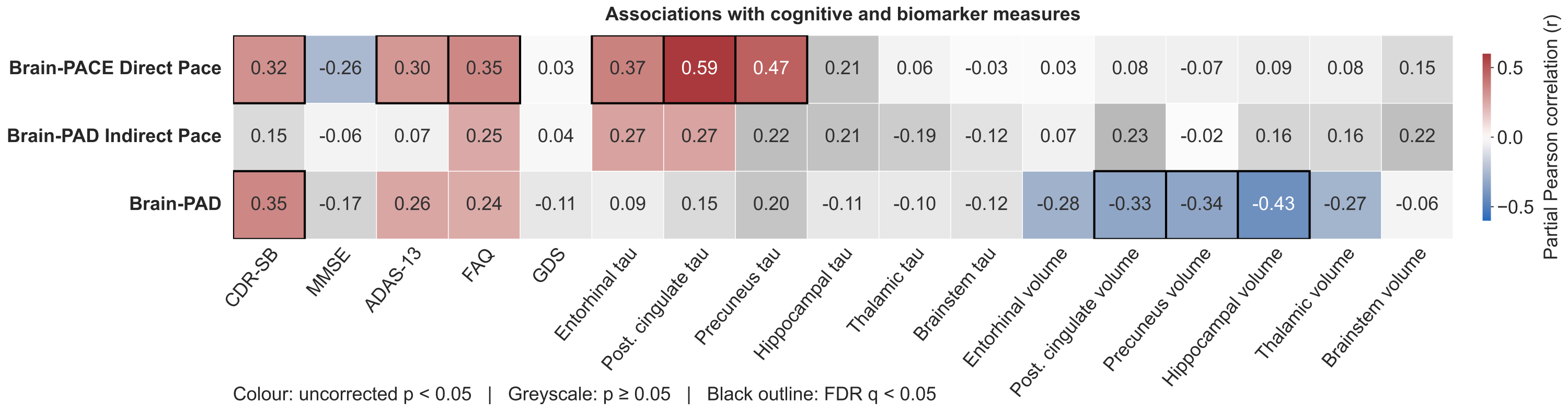}
\caption{Partial correlations between direct Brain-PACE, indirectly derived brain age pace, and cross-sectional brain age gap with cognitive and biomarker measures in the ADNI Mild Cognitive Impairment group. Direct Brain-PACE and indirect brain age pace correlations control for baseline age and median temporal gap; brain age gap correlations control for baseline age. Cells show partial Pearson correlation coefficients ($r$). Coloured cells indicate associations with an uncorrected $p<0.05$, greyscale cells indicate $p\geq0.05$, and black outlines indicate associations surviving false discovery rate correction ($q<0.05$).}
\label{fig:corr}
\end{figure}

For comparison, cross-sectional brain-PAD was primarily associated with structural features and the CDR score. A greater brain-PAD was associated with lower hippocampal volume ($r=-0.43$, $q=0.005$), greater CDR score ($r=0.35$, $q=0.025$), lower precuneus volume ($r=-0.34$, $q=0.025$), and lower posterior cingulate volume ($r=-0.33$, $q=0.025$). In contrast, associations between cross-sectional brain-PAD and regional tau SUVR did not survive FDR correction. Pace derived indirectly from changes in cross-sectional brain age showed weaker associations, with no feature correlation surviving FDR correction. The strongest trends were observed for entorhinal tau SUVR ($r=0.27$, $q=0.201$), posterior cingulate tau SUVR ($r=0.27$, $q=0.201$), and FAQ score ($r=0.25$, $q=0.201$). These results indicate that directly estimated $P$ is more closely coupled to regional tau burden and clinical severity than either cross-sectional brain-PAD or $P$ derived indirectly from combined brain age estimates. The full pattern of associations for Brain-PACE, brain-PAD, and indirect $P$ is shown in Figure~\ref{fig:corr}.

% \FloatBarrier 
\section{Discussion}

Cross-sectional brain age metrics describe how old an individual's brain appears relative to a reference population, but do not necessarily quantify how rapidly that brain is currently changing \citep{cole2017predicting, smith2025characterising}. This distinction is increasingly supported by longitudinal work showing limited agreement between cross-sectional brain age deviations and within-person brain change \citep{vidal2021individual, korbmacher2025cross_hbm}. Building directly on the LILAC and LILAC+ frameworks of \citet{kim2025pace2} and \citet{wegmann2025pace3}, we developed Brain-PACE to estimate longitudinal brain ageing from paired MRI acquisitions. Our central finding is that modelling the differences between time points captures information that is not recovered when pace is derived via repeated cross-sectional brain age estimates. 

Brain-PACE generalised across independent healthy cohorts, identified accelerated ageing in $42.6\%$ of MCI participants, and associated faster ageing with greater cognitive impairment and regional tau burden. These findings extend developing literature showing that MRI derived pace measures capture clinically relevant variation \citep{yin2025pace1, whitman2025dunedinpacni}, while providing a direct comparison against conventional brain age approaches. Methodologically, our contribution focuses on developing the LILAC framework by combining spatial attention with SLD learning using a Cram\'er distance objective and explicit probabilistic uncertainty estimation. Relative to the LILAC+ baseline, Brain-PACE reduced pair-level MAE from $0.779$ to $0.663$ years and MSE from $1.059$ to $0.756$ years$^2$. Cram\'er distance was particularly beneficial for probabilistic performance and prediction bias, while spatial attention provided a complementary mechanism for weighting informative longitudinal features. IG analysis further indicated that predictions were driven by distributed cortical, subcortical, and brainstem changes rather than a single anatomical area. Cram\'er-attention generally improved entropy and empirical coverage relative to KL-attention, although the uncertainty distribution was still less calibrated when structural change becomes pathological. Reduced coverage in MCI and WRAP indicates that uncertainty calibration remains sensitive to cohort and pathological differences. 

Resulting Brain-PACE showed a different biological and clinical profile from both cross-sectional brain-PAD and indirectly derived pace. After normalisation in reference to the control group, $42.6\%$ of MCI participants were classified as showing accelerated Brain-PACE compared with ~$5-10\%$ in the corresponding healthy external test set groups. Using the indirectly derived pace measure, only $20.6\%$ were classified as accelerated in the ADNI MCI group. Faster Brain-PACE was associated with greater functional and cognitive impairment and higher tau burden in the posterior cingulate, precuneus and entorhinal cortex. No statistical significance was observed for these same tau comparisons with either indirectly derived pace or cross-sectional brain-PAD. Brain-PAD did associate with structural measures, including hippocampal, precuneus and posterior cingulate volume. Whilst previous work has highlighted links between brain-PAD and AD pathology \citep{lee2022deep} in our analysis, direct Brain-PACE appeared more closely coupled to the selected measures of active pathological and clinical variation. This is consistent with the wider ideas that an individual's current biological state and their ongoing rate of change can contain separate insights into the same disease processes. 

The association between Brain-PACE and regional tau burden in the MCI group provides particularly important insights for the resulting pace measure. Tau pathology accumulation is closely linked to cognitive decline across the stages of AD \citep{hanseeuw2019association, ossenkoppele2021accuracy, smith2023tau}, and the entorhinal cortex is among the earliest regions affected by neurofibrillary tau pathology \citep{braak1991neuropathological}. The precuneus and posterior cingulate further contribute to understood trajectories of tau propagation \citep{vogel2021four}. Furthermore, tau PET is known to provide stronger prognostic information for subsequent cognitive decline than volumetric measures during preclinical disease stages \citep{ossenkoppele2021accuracy}. These findings suggest that accelerated Brain-PACE may reflect structural change occurring alongside active tau-related neurodegeneration (rather isolated volumetric change). Longitudinal analysis of tau PET is therefore required alongside Brain-PACE to determine if changes are in parallel with observed accumulation and spatial spread of pathology over time. 

The distinction between biological age and its rate of change is also important beyond neuroimaging. Longitudinal changes in epigenetic clocks have recently been shown to predict mortality independently of baseline epigenetic age \citep{kuo2026longitudinal}, providing evidence that temporal trajectories can contain information not represented by a single measurement. Similarly, \citet{liu2026accelerated} derived organ-specific ageing pace from changes in proteomic organ-age gaps over approximately two decades and found that faster ageing of several organ systems, including the brain, was associated with subsequent dementia risk. Both approaches reconstruct pace from repeated cross-sectional age estimates. Our results with Brain-PACE suggest that learning temporal change directly from paired inputs can recover additional information that may be suppressed when biological age acceleration is evaluated indirectly. The methodology introduced here may therefore be relevant beyond structural MRI. Direct paired-input modelling could reduce the propagation of error between repeated age estimates, and distance-aware distributional learning may better represent ordered longitudinal change across various data modalities. Applied to longitudinal epigenetic, proteomic, or multi-organ ageing studies, Brain-PACE could potentially provide a more direct and better calibrated estimate of biological pace.

Future work should evaluate more diverse datasets across different healthcare settings in order to confirm the utility of Brain-PACE. The training and evaluation cohorts did primarily represent older adults, meaning the current model should be interpreted as a measure of later-life brain ageing rather than ageing across the lifespan. Development, midlife ageing and late-life brain ageing may follow different and potentially non-linear structural trajectories so age-specific or lifespan reference models may be required. Additionally, the biomarker analysis was restricted to the ADNI MCI group, so observed associations with tau and clinical severity do not establish whether accelerated Brain-PACE predicts subsequent decline, or is specific to AD-related pathology. Testing cognitively unimpaired individuals with preclinical amyloid or tau pathology, and  assessment of clinical progression and treatment response will be important to establish clinical validity. Finally, Brain-PACE currently uses T1w structural MRI. Ageing is heterogeneous across biological systems and organs \citep{oh2023organ}, and multimodal longitudinal models may better distinguish the processes that contribute to an accelerated pace and determine whether combining complementary markers provides additional downstream utility \citep{mitolo2024association}.

Overall, Brain-PACE extends current brain age modelling by moving from cross-sectional methods to direct estimation of longitudinal structural change. By applying a Siamese architecture with spatial attention and Cramér loss with SLD learning, we improve interval prediction and reduce prediction bias. Associations with tau burden and clinical severity indicate that the resulting $P$ measure captures biologically meaningful variation relevant to early neurodegeneration. To the best of our knowledge, this is the first longitudinal pace framework to be validated against protein-based measures of AD pathology and to apply Cramér distance loss alongside SLD learning to this task. Brain-PACE should not replace brain-PAD, volumetric or disease-specific biomarkers, but act as a complementary longitudinal imaging feature for quantifying ongoing brain health and disease. Future work will evaluate more diverse datasets to better understand if this measure could support subject level risk profiling, preclinical detection and treatment monitoring of AD.

% \FloatBarrier  
\section{Acknowledgements}\label{}
The research presented in this paper was carried out on the High Performance Computing Cluster supported by the Research and Specialist Computing Support service at the University of East Anglia.

Data collection and sharing for the Alzheimer's Disease Neuroimaging Initiative (ADNI) is funded by the National Institute on Aging (National Institutes of Health Grant U19AG024904). The grantee organization is the Northern California Institute for Research and Education. In the past, ADNI has also received funding from the National Institute of Biomedical Imaging and Bioengineering, the Canadian Institutes of Health Research, and private sector contributions through the Foundation for the National Institutes of Health (FNIH) including generous contributions from the following: AbbVie, Alzheimer’s Association; Alzheimer’s Drug Discovery Foundation; Araclon Biotech; BioClinica, Inc.; Biogen; BristolMyers Squibb Company; CereSpir, Inc.; Cogstate; Eisai Inc.; Elan Pharmaceuticals, Inc.; Eli Lilly and Company; EuroImmun; F. Hoffmann-La Roche Ltd and its affiliated company Genentech, Inc.; Fujirebio; GE Healthcare; IXICO Ltd.; Janssen Alzheimer Immunotherapy Research \& Development, LLC.; Johnson \& Johnson Pharmaceutical Research \& Development LLC.; Lumosity; Lundbeck; Merck \& Co., Inc.; Meso Scale Diagnostics, LLC.; NeuroRx Research; Neurotrack Technologies; Novartis Pharmaceuticals Corporation; Pfizer Inc.; Piramal Imaging; Servier; Takeda Pharmaceutical Company; and Transition Therapeutics. Data were provided (in part) by OASIS-3: Longitudinal Multimodal Neuroimaging: Principal Investigators: T. Benzinger, D. Marcus, J. Morris; NIH P30 AG066444, P50 AG00561, P30 NS09857781, P01 AG026276, P01 AG003991, R01 AG043434, UL1 TR000448, R01 EB009352, P30 AG066444, AW00006993. AV-45 doses were provided by Avid Radiopharmaceuticals, a wholly owned subsidiary of Eli Lilly. AV-1451 doses were provided by Avid Radiopharmaceuticals, a wholly owned subsidiary of Eli Lilly. Data were also provided (in part) by WRAP, for which we acknowledge the University of Wisconsin. WRAP is supported by the National Institutes of Health under awards R01AG027161, R01AG021155, R01AG062285, and R01AG062167.

For the purpose of open access, the author has applied a CC BY public copyright licence to any Author Accepted Manuscript version arising from this submission. 

\bibliographystyle{plainnat}
\bibliography{refs}

\end{document}